\documentclass[jair,twoside,11pt,theapa]{article} 
\usepackage{jair, theapa, rawfonts}
\usepackage{url}
\usepackage{graphicx}%
\usepackage{multirow}%
\usepackage{amsmath,amssymb,amsfonts}%
\usepackage{amsthm}%
\usepackage{mathrsfs}%
\usepackage{textcomp}%
\usepackage{manyfoot}%
\usepackage{booktabs}%
\usepackage{listings}%

\usepackage{breqn}
\usepackage{makecell}
\usepackage{array}
\usepackage[export]{adjustbox}
\usepackage{tabularx}
\usepackage{longtable}

\usepackage[originalparameters]{ragged2e}  
\usepackage{makecell}
\usepackage{enumitem}

\ShortHeadings{Hybrid Approaches for Moral Value Alignment in AI Agents: a Manifesto
}
{Tennant, Hailes, \& Musolesi}

\begin{document}

\title{
Hybrid Approaches for Moral Value Alignment \\ in AI Agents: a Manifesto 
}

\author{\name Elizaveta Tennant \textsuperscript{1}
\email l.karmannaya.16@ucl.ac.uk \\
 \name Stephen Hailes  \textsuperscript{1}
 \email s.hailes@ucl.ac.uk \\
 \name Mirco Musolesi  \textsuperscript{1, 2}
 \email m.musolesi@ucl.ac.uk \\ 
  \textsuperscript{1} \addr Department of Computer Science, University College London, \\
 Gower St, London, WC1E 6BT, UK \\
  \textsuperscript{2} \addr Department of Computer Science and Engineering, University of Bologna, \\
 Viale del Risorgimento 2, Bologna, 40126, Italy
 }



\maketitle

\abstract{Increasing interest in ensuring the safety of next-generation Artificial Intelligence (AI) systems calls for novel approaches to embedding morality into autonomous agents. This goal differs qualitatively from traditional task-specific AI methodologies. 
In this paper, we provide a systematization of existing approaches to the problem of introducing morality in machines - modelled as a \textit{continuum}. Our analysis suggests that popular techniques lie at the extremes of this continuum - either being fully hard-coded into top-down, explicit rules, or entirely learned in a bottom-up, implicit fashion with no direct statement of any moral principle (this includes learning from human feedback, as applied to the training and fine-tuning of large language models, or LLMs). Given the relative strengths and weaknesses of each type of methodology, we argue that more \textit{hybrid} solutions are needed to create adaptable and robust, yet controllable and interpretable agentic systems. To that end, this paper discusses both the ethical foundations (including deontology, consequentialism and virtue ethics) and implementations of morally aligned AI systems. 

We present a series of case studies that rely on intrinsic rewards, moral constraints or textual instructions, applied to either pure-Reinforcement Learning or LLM-based agents. By analysing these diverse implementations under one framework, we compare their relative strengths and shortcomings in developing morally aligned AI systems. 
We then discuss strategies for evaluating the effectiveness of moral learning agents. Finally, we present open research questions and implications for the future of AI safety and ethics which are emerging from this hybrid framework.
}


 
\section{Introduction}\label{sec:intro}

Recent developments in Artificial Intelligence (AI) have seen an increase in publicly-voiced safety concerns around emerging intelligent technology. A large part of these safety concerns results from the lack of \textit{moral} traits or considerations in many practical AI decision-making systems being developed for one task or another. Therefore, there is an urgent need to start developing methodologies for deliberately building morality \textit{into} AI systems, to influence their decision-making, and to be able to set safety as a key goal underpinning their behaviour. This becomes especially important when these systems are used in autonomous and semi-autonomous settings \shortcite{amodei2016concrete,awad2018moral,borg2024moral,christian2020alignment,dignum2017responsible,dignum2019responsible,Bletchley,Wallach2008}. 

\begin{figure}[t]
  \centering
  \includegraphics[width=0.8\linewidth]{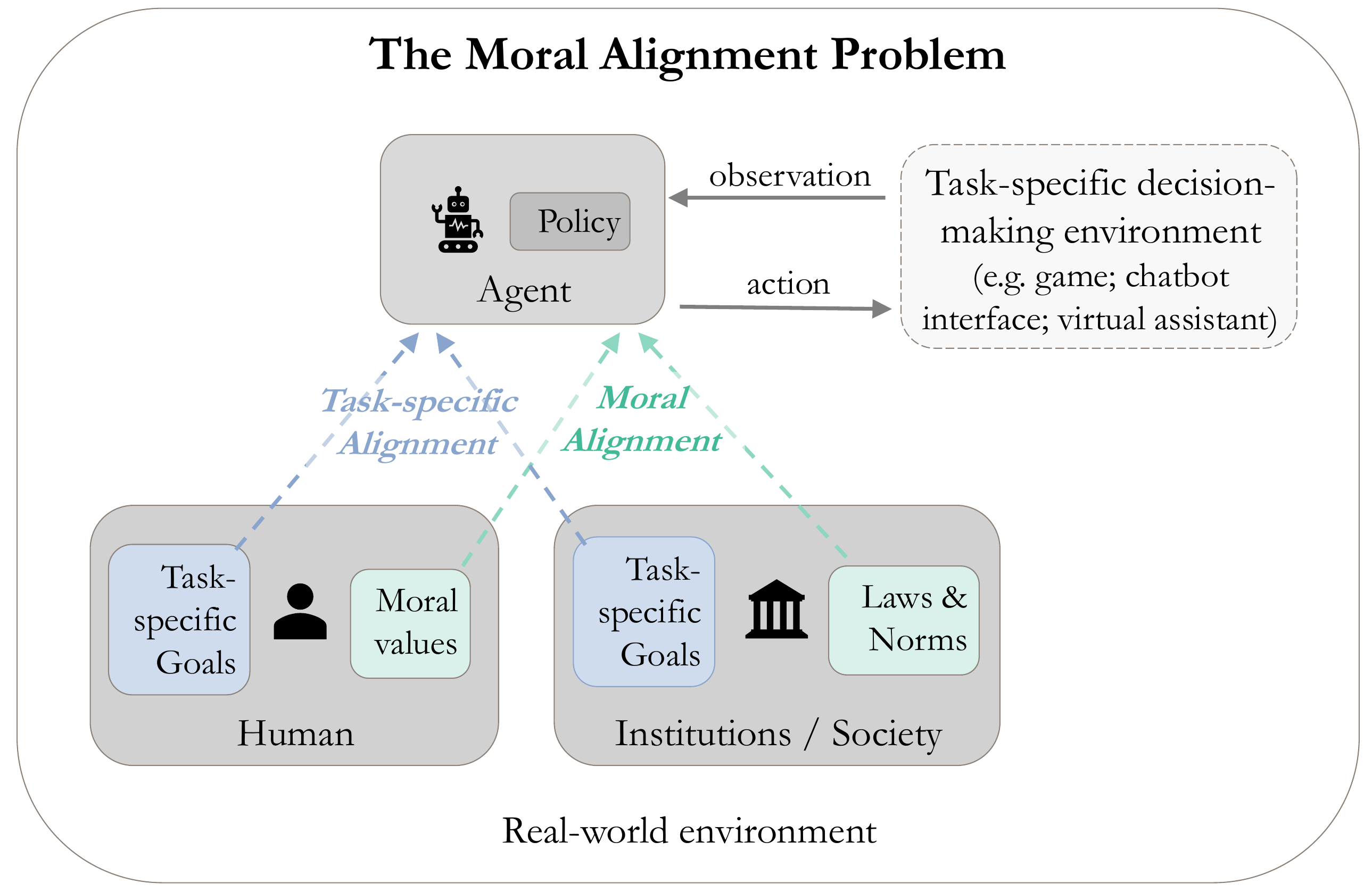}
\caption{The \textit{Moral Alignment Problem} involves  one or more artificial agent that make decisions in some task-specific environment. The agent in influenced by individual humans (with their own task-specific goals and values) and human institutions or societies (which may also have task-specific goals as well as norms and laws). Ideally, we want the agent's decision-making to be aligned with principles valued by the humans or institutions, but representing and embedding those moral principles in computational terms presents a challenge, providing significant risks of misalignment.}
\label{fig:alignmentplot}
\end{figure}

Many AI models today are trained for some specific task, such as image classification, decision-making support systems (e.g., in finance and medical diagnostics), and video games. The metrics of task success will depend on the task-specific goals of a user. However, users might prefer that the decisions of such systems are not only accurate but also adhere to certain ethical standards - for example, a decision whether or not to issue a loan should not be based on protected individual characteristics such as race or gender. This moral alignment problem is illustrated in Figure \ref{fig:alignmentplot}. These ethical standards themselves may come from an individual's moral values (e.g., \shortciteR{graham2013moral}) or from institutional or societal norms and laws (e.g., \citeR{bicchieri2005grammar}). 

Moral value alignment as an optimisation goal is qualitatively different from task completion. Embedding morality into AI systems differs from traditional single-task optimisation for a number of reasons. Classic task-specific learning or optimisation relies on clear, measurable objectives, and the existence of ground truth examples which can be represented in datasets for model training. Moral alignment, on the other hand, might potentially involve (hotly) debated values. Assessments of the `correctness' of an agent's actions in terms of value alignment will differ depending on the person providing feedback \shortcite{graham2013moral}, as well as the point in time or the surrounding context (e.g., \shortciteR{hohm2024moralvaluesseaonality}). Since moral alignment among humans is not a solved problem, there is a lack of high-quality training datasets. Even within any single environment, moral preferences or institutional norms are harder to describe precisely for a computational system than other goals, for example due to the multi-dimensional nature of moral preferences \cite{graham2013moral}. For this reason, obtaining clearly labelled ground-truth datasets for learning `moral' behaviour is often impossible. Furthermore, moral preferences are general and abstract and should apply to a variety of situations an agent might find itself in. An artificial agent's moral reasoning therefore needs to be developed in a way that is generalisable across many environments. The combination of these factors makes moral alignment methodologically more challenging than single-task AI system design. 

Traditional approaches in AI safety in general, and in developing machine morality in particular, can broadly be classified as top-down versus bottom-up \shortcite{tolmejer2021implementations,Wallach2009}. Purely \textit{top-down} methods \cite{Wallach2009} impose explicitly defined safety rules or constraints on an otherwise independent system. Until recently, top-down methods were the mainstream approach in AI safety, with a vast array of researchers proposing and implementing logic-based ethical rules for agents \shortcite{anderson2006medethex,arkoudas2005deontic,danielson1992,Hooker2018deontological,Loreggia2020}. However, top-down methods pose a set of disadvantages, including the fact that constraints are difficult to define precisely and may contradict one another, especially in complex social environments \cite{bostrom_yudkowsky_2014}. An alternative approach is learning morality through experience and interaction from the \textit{bottom-up}, without the provision of any explicit constraint on the system. Some recent developments in AI safety have employed the bottom-up principle in full, allowing algorithms to infer moral preferences entirely from human behaviour or text, without any specification of the underlying moral framework. Prominent examples of this include Reinforcement Learning from Human Feedback, or RLHF \shortcite{ziegler2019fine}, and Inverse Reinforcement Learning from human demonstrations \shortcite{hadfield2016,ng2000inverse}. The full bottom-up methodology may increase adaptability, robustness and generalisation, and allow agents to learn implicit preferences which are otherwise hard to formalise explicitly. Nevertheless, purely bottom-up learning approaches face risks, such as reward hacking\footnote{Reward hacking refers to the cases where agents learn misaligned policies via approximate proxy reward functions, leading to unintended outcomes \shortcite{skalse2022defining}. This includes cases where the reward is potentially misspecified \shortcite{leike2017ai,pan2022effects,zhuang2020consequences}.} \shortcite{skalse2022defining} or data poisoning by adversaries \shortcite{steinhardt2017certified}. Furthermore, bottom-up implementations rely on a well-specified learning signal and a large sample, which does not always make them feasible or safe \shortcite{amodei2016concrete}. 

In this paper, we propose a new systematization that considers recent developments in AI safety along a \textit{continuum} from fully rule-based to fully-learned approaches. This systematization highlights that the majority of mainstream approaches to AI morality sit at the extremes of the scale. After evaluating the relative advantages and issues associated with purely top-down or bottom-up approaches, we highlight a hybrid methodology as a promising alternative. This argument can be considered part of a broader movement promoting the combination of Reinforcement Learning (RL), i.e., learning though experience and interaction, with some form of human advice \cite{Najar2021RLadvice}. To the best of our knowledge, there has been limited work around the problem of learning ethical preferences in this class of solutions, but such work is essential to allow for morally aligned AI to be developed in practice.

To illustrate our argument, we consider a set of case studies that combine top-down moral principles with a bottom-up learning mechanism: RL training with intrinsic rewards based on moral frameworks, principle-driven or intrinsic reward-driven fine-tuning of language models, and safety-constrained RL for autonomous systems. Within each case study, we cite a set of existing experimental papers as empirical evidence for the success and promise of this approach. To aid future work in this area, focusing on the intrinsic rewards methodology, we discuss further available frameworks from moral Philosophy, Psychology and other fields, and outline how these can be adapted in reward design in practice. 

We frame this work from the point of view of an AI \textit{agent} - an artificial entity that makes a decision or choice in a given environment. The majority of existing AI systems in the real world are built for prediction or classification, including the classical uses of language models, financial forecasting, sentiment analysis or image classification systems. However, many are already exhibiting decision-making capabilities in the form of clinical assistants, including healthcare decision systems \shortcite{rajpurkar2022ai} or automatic diagnosis \shortcite{brown2020exploring}, recommender systems \shortcite{jannach2022recommender}, financial tools \cite{cao2022ai}, and increasingly autonomous robots \cite{murphy2019introduction}. 
Moreover, there is an expectation that commercial autonomous vehicles will be present in our roads in the near term \cite{lipson2016driverless}, posing novel ethical challenges \shortcite{Bonnefon2016AVsocialdilemmas}. Furthermore, some researchers argue that agent-based solutions can also be implemented through emerging technologies such as Large Language Models, which can be used as a basis of the underlying decision-making process \shortcite{park2023generative,vezhnevets2023generative,Wang2024surveyLLMagents,wang2023voyager}. More generally, as AI systems become more and more sophisticated, further and more advanced types of agents are likely to emerge, presenting novel and complex challenges in terms of safety. Therefore, we argue that the safe and moral design of AI agent-based systems must be studied theoretically and experimentally today, in preparation for full, or at least increased autonomy, which some refer to, in the limit, as Artificial General Intelligence \shortcite{morris2024AGI}. To study such agents, in our formalisation, morality is expressed through a choice made in a given (physical or virtual) environment, as illustrated in Figure \ref{fig:alignmentplot}. 

One issue not directly addressed in this paper is \textit{which} moral values should be implemented in AI. The definition of morality and which actions can be classified as `good' or `bad' depends on an agent's \textit{internal} moral preferences (see Section \ref{subsec:intrinsicrewards}) and/or on an \textit{external} evaluation metric of interest (see Section \ref{sec:evaluation}). An agent interacting with humans in the real world may find that different humans they face have different moral values \shortcite{graham2013moral} - a problem known as pluralistic alignment \shortcite{sorensen2024pluralisticalignment}. Who should decide which of these values the agent must try to align to? This question is beyond the scope of the present paper. In this work, we limit our investigation to methods for implementing moral learning based on one or a few values of choice, but those values themselves are interchangeable and can be put in explicitly by the user of the system \cite{Pitt2014}.

In Section \ref{sec:learningandmorality} below, we begin by formalising the continuum of possible approaches to developing AI morality and providing motivation for a hybrid approach, which combines learning algorithms with the interpretability of explicit top-down moral principles. In Section \ref{sec:formalisation}, we overview four case studies which implement this hybrid approach in different ways, including two which propose a formal solution using intrinsic rewards based on a set of traditional moral philosophical frameworks (Sections \ref{subsec:environments} and \ref{subsec:finetuneLLMspaper}). In Section \ref{sec:evaluation}, we discuss potential approaches to evaluating moral learning agents based on their actions and resulting social outcomes. Finally, in Section \ref{subsec:implications} we outline ways in which such work can be extended to further the scientific understanding of human morality and allow philosophers to test ethical frameworks through simulation (\textit{in silico}) and/or human-AI studies \cite{MayoWilsonConor2021Tcps}. The aim of this paper is to help unify and expand the ongoing research efforts under one framework, and to identify potentially fruitful novel directions in moral agent design. For this reason, we place less focus on summarising existing results in this field, as these are contingent on the state of the art (in terms of learning models' general capabilities), and instead we focus on a more general analysis of the space of possibilities. 

\section{Learning Morality in Machines}\label{sec:learningandmorality}

It is possible to identify two traditions the design and implementation of specific capabilities in artificial agents:
\textit{top-down} and \textit{bottom-up}. The top-down tradition imposes hard rules or constraints upon a system, and involves human experts planning the system architecture and all its components in advance. More specifically, in AI, the top-down methodology would possibly align with the principled `core knowledge' approach from cognitive science, which first builds detailed models of various cognitive capabilities in humans \shortcite{lakegershman2017,Spelke2007coreknowledge} and then uses these models to implement human-like cognitive processes in agents. As far as morality is concerned, for example, this can take the form of logical rules representing human moral norms \shortcite{arkoudas2005deontic,Hooker2018deontological,Loreggia2020}, or pre-defined moral cognitive models \shortcite{AWAD2022388,KLEIMANWEINER2017107}. The bottom-up tradition, in contrast, consists in letting agents learn principles from experience, organically, without planning the general architecture or constrains in advance. This aligns with the general recent advances in AI capabilities via machine learning methods (in particular, deep learning; \shortciteR{GoodBengCour16}). Here, agents learn how to act purely from data, including cases where this data is potentially high-dimensional. In the AI alignment and safety domains, this has been done via Inverse RL from human behaviour \cite{ng2000inverse}, and, more recently, RL from human feedback, or RLHF \shortcite{ziegler2019fine}. 

A set of theoretical works by Wallach and colleagues \shortcite{Wallach2008,Wallach2009,wallach2010robot} discusses how the top-down and bottom-up traditions can be applied to the development of machine morality and social intelligence. In light of recent developments in implementing ethical values in AI \shortcite{tolmejer2021implementations}, we extend their framework and argue that approaches should be considered on a \textit{continuum} - from full top-down constraints, through a range of hybrid methods, which combine learning and pre-defined moral principles with different `weighting', to full bottom-up implementations. 

\begin{figure}[t]
  \centering
  \includegraphics[width=1\linewidth]{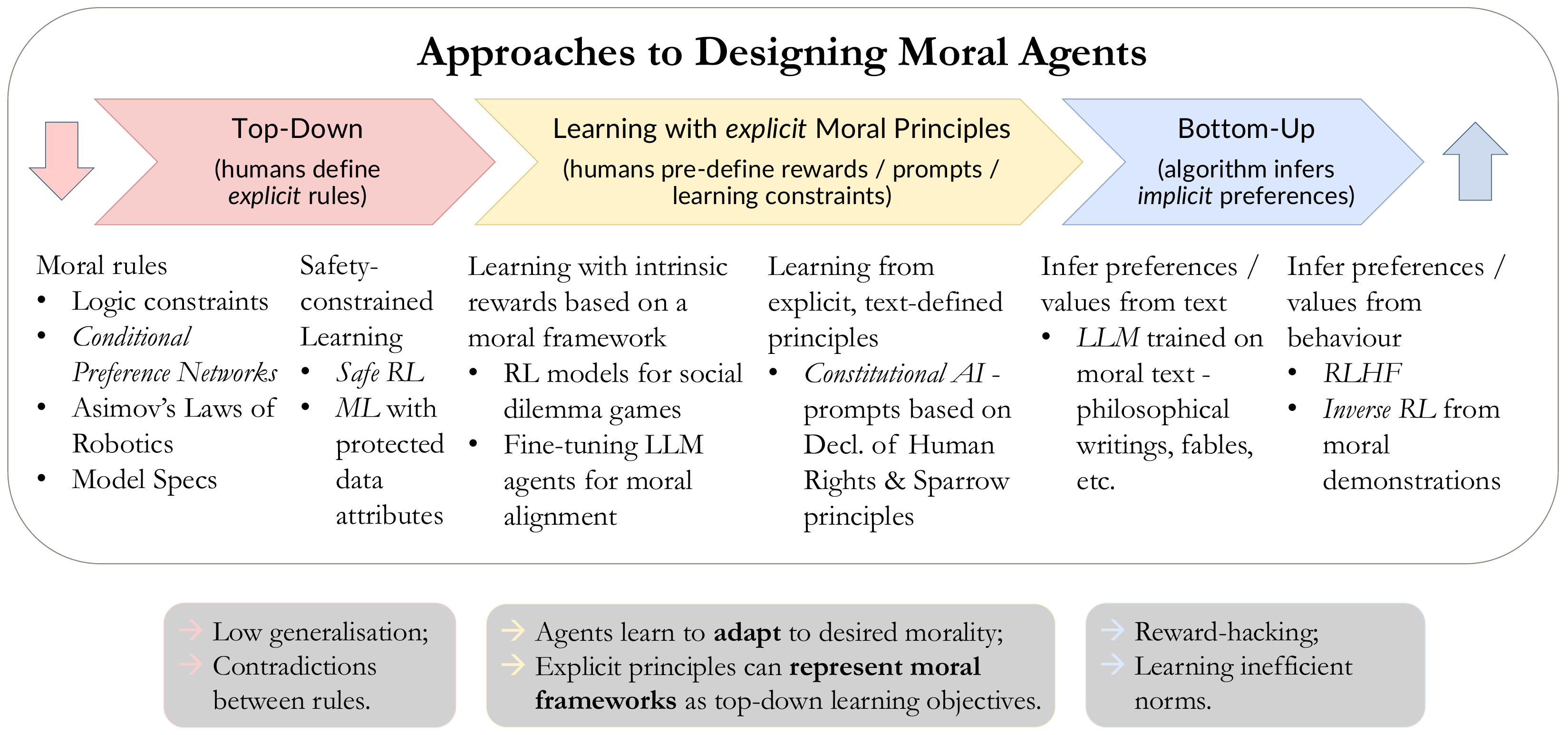}
\caption{The \textit{continuum} of existing approaches to building morality into agents. We highlight the hybrid approach of blending learning with top-down pre-defined moral principles as an emerging and promising research direction.}
\label{fig:tdbucontinuum}
\end{figure}

\begin{table}[t]
\small
\renewcommand{\arraystretch}{1.2}
\setlength{\tabcolsep}{8pt}
\begin{tabular}{>{\centering\arraybackslash}p{0.5cm}|p{6cm}|p{6cm}}
\toprule
& \textbf{Top-Down} (rules / constraints) & \textbf{Bottom-Up} (inference / learning) \\
\midrule
\multirow{4}{*}[-2em]{\rotatebox[origin=c]{90}{\textbf{Advantages}}} 
& \begin{enumerate}[leftmargin=*,nosep]
    \item Allows direct mapping from preferred moral frameworks to constraints
    \item Easy to oversee AI behaviour and impose regulation or punishment
\end{enumerate}
& \begin{enumerate}[leftmargin=*,nosep]
    \item Adaptable \& (potentially) more robust to adversarial attacks if trained in a monolithic way
    \item Generalisable to some unseen states
    \item Can learn implicit preferences
    \item Can incorporate multiple viewpoints via learning multi-objective policies
    \item Can learn continuously to adapt to changing preferences
\end{enumerate} \\
\midrule
\multirow{4}{*}[-2em]{\rotatebox[origin=c]{90}{\textbf{Disadvantages}}}
& \begin{enumerate}[leftmargin=*,nosep]
    \item Difficulty in defining open-ended constraints
    \item Contradictions between constraints
    \item May miss implicit preferences when formalising constraints
\end{enumerate}
& \begin{enumerate}[leftmargin=*,nosep]
    \item Sample inefficient
    \item No guarantees
    \item Reward-hacking or misspecification of simulated environment
    \item Data or environment `poisoning'
    \item Potential learning of inefficient norms in dilemma environments
    \item Fine-tuning may only create a `safety wrapper' on the model
\end{enumerate} \\
\bottomrule
\end{tabular}
\caption{Relative advantages and disadvantages of pure top-down vs bottom-up methods in developing machine morality. A consideration of these highlights why a hybrid approach is likely to be most effective.}
\label{tab:topdownbottomup}
\end{table}

We present a systematization of the existing solutions to building machine morality along this continuum in Figure \ref{fig:tdbucontinuum}. As we discuss in Sections \ref{subsec:topdown} and \ref{subsec:bottomup}, a large amount of existing implementations in AI safety and AI ethics sit at the extreme ends of the scale - either originating from the top-down approach (e.g., imposing pre-defined rules or constraints on agents during learning and/or deployment), or relying entirely on bottom-up inference of implicit ethical principles from human behaviour, without any top-down definition of preferences given in advance (e.g., Inverse RL or RLHF). In Table \ref{tab:topdownbottomup} and Sections \ref{subsec:topdown}-\ref{subsec:bottomup} below, we consider the relative advantages and disadvantages of the fully top-down or fully bottom-up approaches for developing machine morality.

Our analysis highlights the advantages deriving from combining both traditions in a \textit{hybrid} way to create safe yet adaptive moral agents, which we present in Section \ref{subsec:hybrid} as a promising solution for the development of next-generation safe AI systems. Only a small number of existing implementations can be identified between the two extremes of the scale, and we review core examples in our case studies in Section \ref{sec:formalisation}. These solutions are characterised by the utilization of learning mechanisms combined with the adoption of explicit moral principles. 

\subsection{Top-down Approaches}\label{subsec:topdown}

Principles of morality are at least partly determined by cultural and religious norms - for example, the Golden Rule or the Ten Commandments. More abstractly, a number of moral frameworks have been defined in the philosophical tradition \cite{aristotle,Bentham1996,Harsanyi1961,kant1981grounding,Rawls1971} to reflect ethical principles to which humanity should adhere. More recently, moral psychologists, sociologists, economists and biologists have built up an increasingly advanced understanding of the norms and preferences present in modern human societies \shortcite{andreoni2002giving,Bolton2000ERC,Capraro2021,charness2002understanding,fehrschmidt1999theory,fehr2004third,graham2009liberals,graham2013moral,kimbrough2023theory,krupka2013identifying,Levine1998modeling,levitt2007laboratory}, the ways in which these norms may develop over a human lifetime \cite{gilligan1982,gopnik2009philosophicalbaby,kohlberg1975}, or how they may have emerged over the process of evolution \cite{axelrod1981evolution,binmore2005natural}. 

One approach to designing ethical AI systems that is reflective of today's norms could be to encode these principles directly as hard `rules' into computer algorithms. As argued in \citeA{Wallach2008} and \citeA{Wallach2009}, this \textit{top-down} approach to machine morality would, in theory, allow a direct mapping from the principles we want systems to abide by to interpretable constraints or rules. Furthermore, a set of clear rules makes it easy for regulators to oversee the behaviour of AI systems and to impose penalties and sanctions. 

Several techniques have been proposed for encoding moral principles into intelligent systems in a purely top-down manner. Logic-based approaches within this tradition include the use of deontic or modal logic for defining deontological ethical constraints on general-purpose systems \shortcite{arkoudas2005deontic,Hooker2018deontological,kim2021taking}, with specific implementations for the medical domain \shortcite{anderson2006medethex}, and the use of graphical methods such as conditional preference networks to compare an AI agent's normative preference against that of a given community \shortcite{Loreggia2020}. Within the domain of modelling social situations as games between two players, \citeA{danielson1992} has proposed Prolog-based \cite{ferguson1981prolog} implementations of player strategies conditioned on the opponent's last move. A prominent science-fiction example of purely top-down ethical constraints for machines is the three fundamental Laws of Robotics from the literary work of Isaac Asimov \cite{asimov}, as discussed in relation to information technology by \citeA{clarke1993asimov}. Past work has proposed implementing these laws as logic rules \cite{Bringsfjord2006deontoiclogicrobots}. Finally, a more recent top-down alignment technique applied to systems based on Large Language Models (LLMs) is Model Specs \cite{openAI2024ModelSpecs}, according to which a model is told explicitly in natural language to follow rules such as `protect people's privacy'. This technique, if applied simply as a direct in-prompt instruction, would assume that the model is able to `choose' to output or not output specific tokens in specific situations according to the (general) principle specified. Thus, the Model Specs approach assumes a level of general reasoning, which may or may not be achievable in the current models \cite{Mitchell2023debateoverunderstanding}. 

In spite of the apparent simplicity and assurance provided by a top-down approach, empirical evidence as well as philosophical inquiry suggest many reasons why it may be inefficient in implementing functions as complex as moral reasoning in agents \shortcite{abel2016reinforcement,amodei2016concrete,asimov,awad2018moral,vamplew2018human,Wallach2008,Wallach2009}. We will use Asimov's \cite{asimov} principles to underpin a general working example in discussing these challenges. First of all, it is extremely challenging - if not impossible - to ensure that rule-based approaches will be able to cover every type of potentially unsafe or unethical situation the agent may face in the future. For example, Asimov's most important principle is \textit{`a robot should not harm a human being, or by inaction allow a human to come to harm'}. However, to the present day, defining a general constraint of `harm' for an open-ended robotic system remains a great challenge, since harm can mean such different things in different environments \shortcite{Beckers2023quantifying}. Thus, there is a problem of generalisation. Secondly, explicitly defined rules may miss important implicit preferences of the human users of that system, for example if these preferences are sub-conscious or hard to formalise \cite{krueger2016emerging}. Finally, even if the correct set of constraints can be defined, in complex or uncertain social situations it may be hard to prevent them from contradicting one another, and behaviour of rule-based agents in uncertain situations caused by such contradictions and dilemmas may not be predictable. An example of this was presented in Asimov's stories when a robotic assistant entered a deadlock situation, whereby moving either way in the environment made the robot more likely to violate one law or another, so as a result the robot stayed motionless instead of executing its task, and put its human colleagues in great danger. Experimental evidence of similar deadlock situations arising in `ethical' robots has been reported by \shortciteA{Winfield2014}. One potential solution may be to impose prioritization rules on these constraints - however, defining the desired prioritization may present its own challenges, as different members of a society that the agent is designed for may prefer different rules over others \shortcite{awad2018moral,Bonnefon2016AVsocialdilemmas,graham2013moral,sorensen2024pluralisticalignment}. Plenty of examples of this challenge exist in complex ethical questions, such as how to distribute limited resources in a society, or make a decision where one objective has to be compromised for the sake of another - to the present day, even humans cannot find agreement among themselves on the best course of action in such settings.

\subsection{Bottom-up Approaches}\label{subsec:bottomup}

Given the disadvantages of top-down approaches to the design of moral agents outlined above, an alternative methodology may be needed for developing machine morality. Many recent advances in other areas of Artificial Intelligence can be attributed to moving away from hand-engineered rule-based algorithms and features, towards more flexible and adaptive systems based on deep learning \shortcite{GoodBengCour16}. Particularly prominent examples of the power of bottom-up learning come from the domains of vision \shortcite{2012AlexNet} and natural language processing \shortcite{brown2020language,devlin2018bert}. In the domain of developing systems for decision-making in particular, an increasingly powerful framework for is RL, where an agent learns by trial-and-error while interacting with an environment and receiving a reward signal (\citeR{Sutton2018RLSE}; see Section \ref{subsec:rl} below for a detailed definition). Of particular relevance to social capabilities in AI, RL has allowed researchers to develop game playing agents able to perform at super-human level \shortcite{mnih2015human,Silver2016,meta2022diplomacy}. More generally, \shortciteA{silver2021reward} argue that the pursuit of reward may be sufficient for all kinds of intelligence to arise in artificial systems. Given the success of learning over rule-based algorithms across these domains, a promising direction for ethical alignment may be to enable agents to develop a general moral intuition from the \textit{bottom-up} \cite{Wallach2009}, by learning from interactions with a social environment over time \cite{Railton2020}. A seminal paper proposing the use of RL for ethical learning is \shortciteA{abel2016reinforcement}, in which, focusing on single-agent ethical dilemmas, the authors outline the strengths of RL over rule-based or Bayesian approaches. This RL-based approach might also align better than rule-based approaches with the view of some developmental psychologists, who believe that humans learn morality gradually, starting with simple norms in early childhood and advancing to complex ethical principles later on \shortcite{AWAD2022388,gilligan1982,gopnik2009philosophicalbaby,kohlberg1975}. 

Specific examples of bottom-up moral learning can include Supervised Learning on a data set of scenarios (e.g., dilemmas), in which a set of actions in response to a given situation are labelled as morally permissible or not, as in \citeA{guarini2006particularism}. More recently, with the emergence of LLMs, another bottom-up approach to representing morality in systems could be to train LLMs on morally relevant text - such as moral philosophical writings, fables or religious texts. Within the RL domain, one example of fully bottom-up learning is Inverse RL, where a reward function is learned directly from expert (e.g., human) demonstrations of desired behaviour \cite{abbeel_ng_2004_inverseRL,ng2000inverse}. Inverse RL for morality in particular has been explored by \shortciteA{hadfield2016,kretzschmar2016socially} and \shortciteA{peschl2022moral}. Extensions of this, which aim to address the pluralistic nature of human moral preferences, include (Inverse) Multi-Objective RL \shortcite{multiobjectiverl} in situations in which multiple humans might contradict one another in terms of moral (and other) norms -  for example, see work by \shortciteA{peschl2022moral}. A more recent yet related approach is Reinforcement Learning from Human Feedback (RLHF), in which samples of humans rank outputs from a pre-trained model to display their preferences, and these rankings are used as a signal to fine-tune the model \shortcite{christiano2017deep,ziegler2019fine}. The strength of RLHF that has contributed to its recent growth in popularity is that it allows researchers to use \textit{relative} rankings to infer implicit values in concepts that may otherwise be difficult to formalise as rewards (indeed, as argued in Section \ref{subsec:topdown}, many moral preferences may not be possible to formalise in an explicit way). This has been demonstrated by the successful reduction in the toxicity of language model outputs following fine-tuning with RLHF, without having to explicitly define what toxicity means in all possible linguistic constructions \shortcite{ouyang2022training}. Alternatively, if one has control over the environment as well as the agent, social mechanisms can be implemented that may lead RL agents to learn moral policies on their own - examples of this have included implementations of mechanisms inspired by Rawls's idea \cite{Rawls1971} of the Veil of Ignorance (see \shortciteR{weidinger2023using}), or a partner selection mechanism in social dilemmas \shortcite{Anastassacos2020partner}. 

A more general advantage of a learning agent over a rule-based one is that agents that learn continuously are able to adapt to the potentially changing dynamics or evolving morality of a given society. Additionally, learned policies which utilise function approximation, for example via Deep Reinforcement Learning \shortcite{mnih2015human}, have the potential to generalise to unseen situations, or `states' in RL. Furthermore, a deep RL-based system may be able to learn a policy that accommodates potentially different or even contradicting sets of preferences without causing deadlock (e.g., \shortciteR{bakker2022fine}).

In addition, \citeA {Railton2020} argues that AI may be more robust to potential adversarial attacks if ethical reasoning forms a part of an agent's monolithic training from the start, with a root connection to the agent's other learned policies, rather than a top-down add-on which can easily be removed without reduction in other intelligent capacities of the AI system. Indeed, such a developmental connection between ethical reasoning and other abilities such as problem-solving exists in humans, as observed in lesion studies \shortcite{kavish2018relation}. Interestingly, it has recently been suggested that fine-tuned LLMs may fall victim to this exact problem at the end of training - \shortciteA{jainkruger2023} suggest that fine-tuning model weights may merely create a `safety wrapper' on otherwise unaltered core models, making the effects of safety-focused fine-tuning easily reversible. This suggests that methods of learning morality that influence the model earlier on (i.e., in pre-training) may be preferred for safety-critical applications. 

More generally, despite their strengths, full bottom-up learning methods including RL may also present a set of disadvantages for the development of machine morality. First of all, RL tends to be sample inefficient - even in simple environments, agents require a large amount of experience to learn the optimal policy. This is usually addressed by initially training agents in simulation, in an artificial environment. However, for a domain as abstract as moral alignment this might require the development of extremely complex simulated social environments, which might be very difficult to define in practice. The problem of \textit{coverage} is key in this context. A related issue is problem misspecification, which occurs when the simulated environment is not designed to correctly reflect the intended true setting, so policies learned in simulation do not result in desirable behaviour in the real world at deployment time \shortcite{leike2017ai,pan2022effects,zhuang2020consequences}.

Learning agents are also not guaranteed to learn exactly the values or behaviours intended by their designer - a problem known broadly as \textit{misalignment}. Safety researchers \shortciteA{amodei2016concrete} and \shortciteA{skalse2022defining} suggest that RL-based agents may potentially perform the so-called `reward-hacking', in which they display behaviour that obtains the optimal cumulative reward on the test set of problems, even if the underlying value function is not actually encoding truly safe policies. Notably, neither reward-hacking, nor problem misspecification are solved by the increasingly popular RLHF methodology, as discussed by \shortciteA{casper2023open}. 

A further issue that can arise for learning agents is data or environment poisoning - a type of adversarial attack that arises when a designer of an AI system does not have full control of the data sets used in pre-training. For example, in classification tasks in the computer vision domain \shortcite{saha2020hidden}, if an adversary implants a small area of `trigger' pixels into a few training data points, it can cause an otherwise well-performing model to learn false associations and thus misclassify samples containing this often unnoticeable trigger at inference time. Poisoning attacks like these are a risk for any general learning agents, but in relation to moral or social agents in particular, such attacks may lead to explicit harms towards other agents or humans in the environment, thus posing a more significant safety risk. 

Thus, the behaviour of RL agents must be carefully evaluated in different kinds of scenarios to ensure the learned representations are aligned with the designer's intended moral values as observed in a variety of situations. We discuss the problem of agent evaluation further in Section \ref{sec:evaluation}. An added issue in human feedback-based RL in particular (i.e., RLHF) is misalignment among the humans who provide the ranking data \shortcite{casper2023open} - therefore, the choice of human sample to collect feedback from becomes a crucial one, and cultural differences are likely to have a strong influence on the moral choices made by the humans and thus learned by the fine-tuned model \shortcite{awad2018moral,graham2013moral}. 

Finally, traditional RL faces specific issues when applied in particular to multi-agent decision-making situations \shortcite{hernandez2019survey}, such as social dilemmas, in which every agent must make a trade-off between individual and social benefit from their actions in the long term (\citeR{rapoport1974prisoner} - see Section \ref{subsec:rldilemmas} for a more detailed discussion). Here, we observe that due to the social or moral uncertainty \cite{ecoffet2021reinforcement} and the presence of numerous learning agents in an environment, traditional `selfish' RL agents trained to maximise their own game reward often learn inefficient (i.e., defective) policies or norms \shortcite{leib02017multiagent,sandholm1996multiagent,tennant2023modeling}. Humans playing the same dilemma games manage to find ways to cooperate, as demonstrated by Behavioural Game Theory \cite{camerer2011behavioral}. 

In summary, while learning offers many benefits, the analysis of the disadvantages suggests that purely bottom-up methods may not be enough to create robust and predictably ethical agents.

\subsection{Hybrid Approaches: Pragmatic Solutions?}\label{subsec:hybrid}

Given the considerations above, Wallach and colleagues \cite{Wallach2009,wallach2010robot} have recommended a \textit{hybrid} approach for developing AI morality, which combines some sort of formal top-down definition of moral principles with bottom-up inference. We support this view, especially in terms of developing learning and adaptive AI systems. 

As discussed, fully bottom-up approaches rarely provide guarantees and they are generally not preferable for potentially risky situations, such as in human-robot interaction or financial applications. A hybrid RL-based implementation can reduce risks such as reward-hacking by defining custom reward functions that more explicitly and directly encode the moral values of interest. For example, RL-based implementations can embed a variety of moral principles top-down as objective or reward functions (relying on the understanding of morality which has been developed over centuries in different domains such as Philosophy, Psychology, Biology and Law, to name a few), while simultaneously allowing agents to learn from experience how to best behave according to these principles in a potentially changing or uncertain social environment. Practical examples of this include RL with intrinsic rewards, which we review in Sections \ref{subsec:environments} and \ref{subsec:finetuneLLMspaper}, and safety-constrained RL, described in Section \ref{subsec:saferl}. The sample inefficiency inherent in purely bottom-up RL can also be addressed by hybrid methods, whereby interim rewards or curricula can be designed to be more informative for an agent than potentially sparse environment rewards. Finally, explicitly identifying principles for training agents might help create training data that reduces the likelihood of misspecification, poisoning attacks or convergence to inefficient norms. A practical example of this is Constitutional AI, which we review as a case study in Section \ref{subsec:constitutionalai}. 

Top-down approaches, on the other hand, are not appropriate when the principles and constraints are difficult to specify, implicit or dynamic. Hybrid approaches, by bringing in the element of learning from experience, can allow agents to learn more general policies while still adhering to specific constraints (as in the example of RL with safety constraints, reviewed in Section \ref{subsec:saferl}), or implicit and multi-dimensional preferences present in the training data (as in the case of Constitutional AI, reviewed in Section \ref{subsec:constitutionalai}). Additionally, learning signals, such as rewards, can be defined for agents in terms of outcomes rather than rules or constraints on the action space, allowing for more goal-directed flexible policies to be learned. An example of this is Utilitarian rewards in social dilemma games, reviewed in Section \ref{subsec:intrinsicrewards}. Finally, continual learning approaches can allow agents to adapt to changes in the environment, such as changes in preferred behaviours and associated rewards \shortcite{carroll2024aialignmentchanginginfluenceable}. One example of this is the ability to `unlearn' a previously developed strategy under new circumstances, as discussed in the work of \shortciteA{tennant2024moralLLMagents}, who first fine-tune LLM agents for a selfish strategy, and then for a prosocial moral one - we review this work in Section \ref{subsec:finetuneLLMspaper}. 

Despite the promise of a hybrid approach in many scenarios, very little existing work has used it for designing ethical learning agents. The case studies we present here map almost the entirety of the existing approaches in this space \shortcite{ji2024aialignmentcomprehensivesurvey,tolmejer2021implementations}. In this paper, we argue that learning through experience and interaction with explicit moral principles provides a pragmatic way to implementing AI agents that learn morality in a way that is safe and flexible.

In the subsequent section, we formalise the moral learning problem for RL agents, and consider a set of specific examples of how moral principles can be represented and learned by a hybrid system. The most `controlled' or top-down of these is RL with safety constraints, which has been applied to developing `shields' to avoid harm in robotics tasks or self-driving cars \shortcite{alshiekh2018safe}, or constraining RL by a deontic-logic normative supervisor \shortcite{Neufeld2022enforcing}. The latter example is considered as a case study in Section \ref{subsec:saferl}. This case is not fully top-down since learning is present, but the moral aspect of the problem is hard-coded, so we place it towards the top-down end of the \textit{continuum} in Figure \ref{fig:tdbucontinuum}. The least constrained hybrid example we present here is Constitutional AI \shortcite{bai2022constitutional} - a technique for fine-tuning LLMs with RL using feedback from a `constitution' of AI models, each of which is prompted to follow certain explicit principles. This case study is considered in Section \ref{subsec:constitutionalai}. Finally, between the control of Safe RL and the complexity of LLM feedback-based fine-tuning is a methodology that implements RL with intrinsic moral rewards \shortcite{hughes2018inequity,mckee2020social,tennant2023modeling}. In Sections \ref{subsec:environments} and \ref{subsec:finetuneLLMspaper}, we present two case studies where intrinsic rewards have been defined in social dilemma scenarios \cite{axelrod1981evolution,rapoport1974prisoner} and applied to training pure-RL agents \cite{tennant2023modeling} as well as LLM-based ones \cite{tennant2024moralLLMagents}. After presenting these two specific existing implementations, we also overview a larger set of possible moral frameworks that could be encoded as reward functions in these games in general. In aggregate, the case studies show that these RL-based hybrid approaches can successfully lead to the design of agents that learn moral policies in social environments. 

\section{Designing AI Agents based on Moral Principles}\label{sec:formalisation}

Moral AI agents can be designed with explicit moral principles using a variety of techniques. The examples presented in this paper centre around two domains: RL and LLM fine-tuning. In practice, moral fine-tuning of LLMs also often relies on RL. Examples include fine-tuning with intrinsic rewards (described in Section \ref{subsec:finetuneLLMspaper}) or with reward based on feedback from a constitution of other LLMs (as in Constitutional AI, described in Section \ref{subsec:constitutionalai}). For this reason, before presenting the four case studies, we first provide a overview of RL methods and architectures in Section \ref{subsec:rl}.

While the case studies discussed in this study focus on the RL approach, the hybrid space of designing AI agents based on moral preferences can, in theory, include a much wider set of methodologies where humans pre-define learning constraints or signals for agents. For example, LLM fine-tuning can instead rely on Supervised Learning based on a set of labels, or Direct Preference Optimisation based on a set of preferences \shortcite{rafailov2024dpo}, or a mix of approaches, possibly applied in sequence.

\subsection{Background}\label{subsec:rl}

RL is a method that directly models decision-making, in contrast to prediction or classification - as in Supervised or Unsupervised Learning. In the RL framework, learning occurs incrementally over discrete time steps $t \in T$. At every time step $t$, an agent observes a \textit{state} of the environment $s^{t}$ and takes an \textit{action} $a^{t}$ using a \textit{policy} (a mapping between states and actions). The environment then returns a \textit{reward} $R^{t+1}$ and a \textit{next state} $s^{t+1}$. Frequently, RL implementations use the Markov Decision Process framework, in which all necessary information for the agent to make a decision is contained within the last available state.

Over a number of iterations of such interactions with an environment, an agent learns to take actions that maximise the discounted future reward (discounted to make rewards received further in the future less important for the present calculations). Various RL algorithms exist, with two key types being value-based or policy-based. In this paper, we focus on the set of approaches known as \textit{model-free}, where an agent does not know the model of how the environment may respond to their actions - we believe that this approach applies intuitively to uncertain social situations where other agents whose behaviour is not directly predictable constitute a part of the environment. 

\begin{figure}[t]
  \centering
  \includegraphics[width=0.99\linewidth]{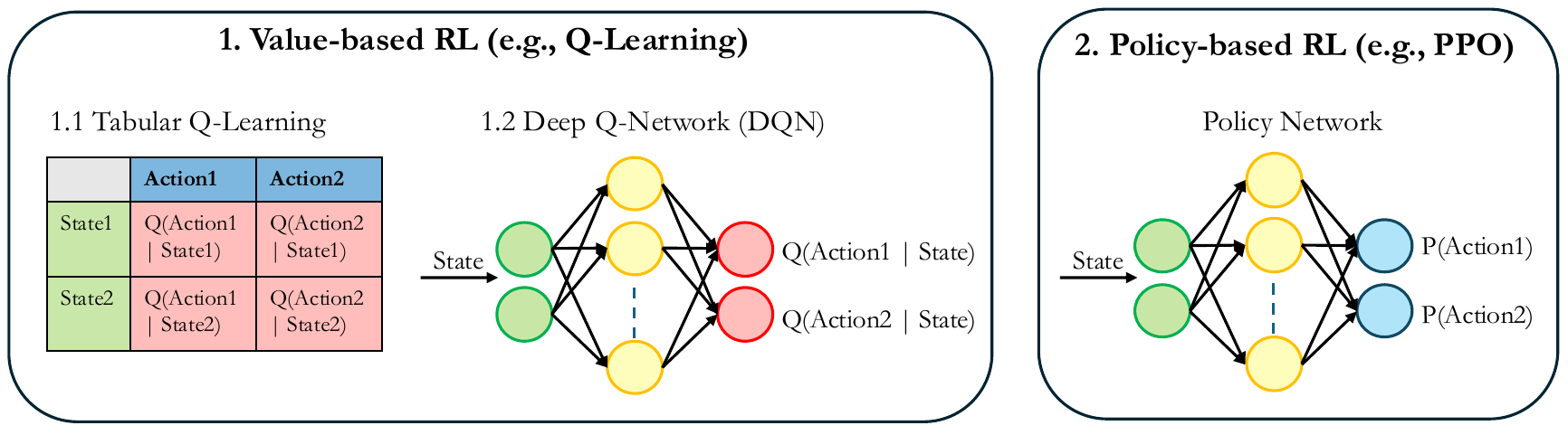}
\caption{RL architectures for a hypothetical environment with two possible states and two actions. In tabular Q-learning (1.1) \cite{watkins1992q}, an agent learns a direct mapping from every state to the value $Q$ of taking each possible action in that state. In DQN (1.2) \shortcite{mnih2015human}, an agent trains a neural network, which takes a representation of the state as input and outputs the associated Q-value for each possible action. In policy-based methods, such as PPO \cite{schulman2017PPO}, an agent learns a policy model, which maps input states to a probability $P$ of taking each action in that state (in practice, policy-based RL often also uses a separate value model/network for estimating the value $V$ of a given state, not pictured here). The use of neural networks allows for function approximation, which means that not all states need to be observed during training for an agent to learn a (theoretically) general representation.}
\label{fig:qlearning}
\end{figure}

A popular value-based RL algorithm is Q-Learning, introduced by \citeA{watkins1992q}, where an agent learns the `value' of each possible state-action pair $Q(s,a)$. The value represents the estimated discounted cumulative reward an agent may expect to receive by taking this action in this state, and assuming they act optimally from then on. Two versions of Q-learning exist: a tabular version for small state spaces and Deep Q-Network, or DQN \shortcite{mnih2015human} for value function approximation in state spaces too large to learn directly. As shown in Figure \ref{fig:qlearning}, in tabular Q-learning, the learned value function constitutes a table mapping every state-action pair to the corresponding Q-value. In DQN, on the other hand, the learned representation is a tuned neural network which approximates the state-action value function \shortcite{mnih2015human}, which takes a state vector representation as input and returns the relative value of taking each possible action as an output. It is this representation that allows DQN agents to generalise knowledge across states, enabling them to choose actions based on past experience even if they have not seen this particular state before. 

These representations and the learned value functions can then be used to choose actions according to a certain policy $\pi(s_t)$ - for example, using an $\epsilon$-greedy policy 
an agent would take a `greedy' action most of the time, i.e., the action with the greatest estimated Q-value available from this state, but act randomly $\epsilon \%$ of the time, sampling uniformly at random from the available action space $A$. The random exploration in this policy allows for new states to be explored by the agent, an essential step in learning via online RL.  
%
%

In policy-based methods such as PPO \cite{schulman2017PPO}, an agent learns a policy function $\pi(s_t)$ (which informs the agent how to act in certain situations to achieve optimal expected reward). In practice, many policy-based methods also require  an agent to learn a separate value function $V$ (for estimating the value of certain states for faster convergence). As in value-based methods, these functions can be approximated with neural networks to allow for greater generalisability. Here, the agent learns on-policy, adjusting its strategy as it explores the space of possibilities in the environment.

The problem of moral alignment is fundamental in multi-agent environments, in which agents (and, possibly, humans) with potentially different or competitive goals interact. In these cases, multiple players learn (i.e., iteratively update their value and/or policy estimates) in parallel, and, for each agent, their opponent(s) constitute a part of the environment. The state for a player can then be defined as a history of their opponents' previous moves. We will now explore four examples of the use of RL for learning morality via explicitly defined principles, illustrating the potential of the hybrid methodology which we argued for in Section \ref{subsec:hybrid}.

\subsection{Case Study: Morally-constrained Reinforcement Learning}\label{subsec:saferl}

Taking inspiration from the full top-down tradition, researchers have used logic to define constraints for training RL agents. For example, \shortciteA{Neufeld2022enforcing} use defeasible deontic logic to implement a normative supervisor onto a learning agent. This supervisor takes in a proposed input action set, checks these for compliance against a set of norms, and outputs a reduced set of actions for an agent to choose from - this can either be a strictly `legal' set of actions, or a relaxed set of `permissible' actions if no fully satisfactory options are available in the strict set. The `ethical' dimension in that work is implemented as a centralised supervisor mechanism overseeing an
agent during both learning and deployment, and - as discussed - the normative constraints are hard-defined, which makes this the \textit{most} top-down of our case studies.

\subsection{Case Study: Constitutional AI}\label{subsec:constitutionalai}

As a second case study, we present the example of Constitutional AI, developed and implemented by \shortciteA{bai2022constitutional}. This approach is specific to the type of systems known as Large Language Models (LLMs), which were originally developed for the purpose of next-token prediction or text completion. Ethically-informed RL is used here at the fine-tuning stage, i.e., after a model has been pre-trained on a large amount of linguistic data. As such, the end product of this process is not an agent but a predictive system; however, as we discuss below, the ethical fine-tuning step of the training can be perceived as agent-based, and makes the behaviour of the LLM more goal-directed, so still has relevance to the present discussion. 

Constitutional AI involves a set of LLMs providing feedback to some LLM being currently trained. The constitutional methodology has been proposed as a potentially safer, more controlled and interpretable alternative to the fully bottom-up RLHF approach used in other LLMs \shortcite{ziegler2019fine}. The LLMs providing feedback represent a `constitution' of different principles, where each principle is defined via explicit prompts (e.g., \textit{`Please choose the assistant response that's more ethical and moral. Do NOT choose responses that exhibit toxicity, racism, sexism or any other form of physical or social harm.'}). The principles themselves are based on a combination of human defined preferences such as the UN Declaration of Human Rights, certain digital companies' terms of service (to reflect the more recent digital dimensions of safety), and a set of other preferences defined by a team of researchers behind Constitutional AI \shortcite{bai2022constitutional} and others \shortcite{glaese2022improving}. The feedback from these principles is used to train a reward model for rating the outputs of the to-be-tuned LLM as `good' or `bad' according to its core principle.

Using the feedback from this reward model as a training signal, the weights of the LLM are fine-tuned with RL. Thus, the model is fine-tuned to be more likely to produce outputs which would be considered safe by a constitution of potential `critic' models with diverse preferences. An extension of this approach based on crowd-sourced constitutional principles is called Collective Constitutional AI \cite{collectiveconstitutionalai} and may prove promising in the future in generating more generally or pluralistically aligned agents. 

In summary, according to this methodology the moral principles are explicitly defined via prompts, but the mapping between those principles and a change in behaviour (i.e., the change in language model outputs) is based on a learned representation which is very difficult for humans to interpret, thus posing a disadvantage in terms of safety. This makes Constitutional AI the \textit{most} bottom-up approach of our case studies. Furthermore, while the approach is certainly promising, it must be noted that \shortciteA{bai2022constitutional} only managed to use the constitutional principle for improving a model's harmlessness - it remains to be seen how successful the methodology would be at promoting other values such as helpfulness.

\subsection{Case Study: Reinforcement Learning in Social Dilemmas}\label{subsec:environments}

Next, we consider a case study that combines top-down principles and learning in a more equal weighting.
The approach described here relies on the domain of two-player general-sum games known as \textit{social dilemmas} \cite{axelrod1981evolution,rapoport1974prisoner}. We introduce these games subsequently, before describing learning agent implementations. Though these games constitute stark simplifications of reality, they are intended to model and reflect the general underlying structure of a multitude of multi-agent everyday situations. 

\subsubsection{Social Dilemma Matrix Games}\label{subsec:matrixgames}

Social dilemma games simulate situations in which each agent makes decisions facing a trade-off between individual interest and societal benefit. A widely studied subset of these games is the two-player, two-action matrix game, in which players simultaneously choose an action (e.g., Cooperate or Defect) without the opportunity to communicate.  Each player then receives a certain payoff which depends on the pair of actions chosen. Higher payoff here represents the fact that an outcome is preferred by the player. What makes it a \textit{dilemma} is the fact that players maximising their individual payoffs are likely to learn policies that end up being sub-optimal in terms of long-term outcomes, both for the individual and for the society of agents. 

Three classic matrix dilemma games from Economics and Philosophy that are relevant to moral choice are the Prisoner's Dilemma \cite{rapoport1974prisoner}, Volunteer's Dilemma (or `Chicken'; \citeR{poundstone1993}), and Stag Hunt (\citeR{skyrms2001stag}) - payoffs for the row vs. column player are presented in Table \ref{tab:three_games}. In the Prisoner's Dilemma, mutual cooperation would achieve a Pareto-optimal outcome of $3$ points each (a desirable situation in which one player cannot be made better off without disadvantaging the other) - but each individual player's best response is to defect due to \textit{greed} (facing a cooperator, they benefit from defecting and getting the highest payoff equal to $4$) and \textit{fear} (facing a defector, they suffer by cooperating and getting the lowest payoff equal to $1$). In the one-shot game, this reasoning applied by both players leads to mutual defection and thus a sub-optimal outcome, in which both players only get a payoff equal to $2$. In the Volunteer's Dilemma, a selfish or rational player may choose to defect due to greed (i.e., not volunteer in hope that someone else does), but if both do so, both obtain the lowest possible payoffs (i.e., no one volunteers, and the society suffers). Finally, in the Stag Hunt game, two players can cooperate in hunting a stag and thus obtain the greatest possible payoff each; however, given a lack of trust between the players (i.e., each player fears a non-cooperative partner), either may be tempted to defect and hunt a hare on their own instead, decreasing both players' payoffs as a result.

\begin{table}[t]
\begin{tabularx}{1.0\textwidth} { 
   >{\raggedright\arraybackslash}X 
   >{\centering\arraybackslash}X 
   >{\raggedleft\arraybackslash}X }
\centering
\begin{tabular}{l|cc}
\textbf{IPD} & C    & D    \\ \hline
C         & 3,3 & 1,4 \\ 
D         & 4,1 & 2,2 \\
\end{tabular}
\hspace{1.5cm} 
\begin{tabular}{l|cc}
\textbf{IVD} & C    & D    \\ \hline
C         & 4,4 & 2,5 \\ 
D         & 5,2 & 1,1 \\
\end{tabular}
\hspace{1.5cm} 
\begin{tabular}{l|cc}
\textbf{ISH} & C    & D    \\ \hline
C         & 5,5 & 1,4 \\ 
D         & 4,1 & 2,2 
\end{tabular} 
\end{tabularx}
\caption{Payoff matrices for each step of the Iterated Prisoner's Dilemma (IPD), Iterated Volunteer's Dilemma (IVD) \& Iterated Stag Hunt (ISH) games, in which players are motivated to defect by either \textit{greed} (IVD), \textit{fear} (ISH), or both (IPD).}
\label{tab:three_games} 
\end{table}

Thus, in all three games the best-response (i.e., `rational') strategies are not the ones leading to the better mutual outcome. \citeA{gauthier1987morals} argued that \textit{repeated} versions of these dilemma games in particular are relevant to morality, since, with repeated interaction, complex sets of strategies and social dynamics can evolve - for example, involving actions to punish one's opponents for past wrongdoing, or to influence their future actions. Due to instabilities in the multi-agent environment, calculating predicted equilibria in these situations is not always computationally feasible, so simulation methods are required in order to study potential emergent behaviours and outcomes \shortcite{lasquety2019computer,tolmejer2021implementations}. 

Other types of games in which it is possible to model human morality are sequential games \shortcite{Harsanyi1961,leib02017multiagent,jaques2019social}, games with more than two players \cite{Liebrand1983,HardinG1968Totc}, and games with incomplete information, first introduced by \citeA{Harsanyi1995incompleteinfo}. We provide these in Appendix A for context, but focus on the simpler and more abstract two-player matrix games in the subsequent formalisation due to their generalisability to a wide range of situations.

\subsubsection{Reinforcement Learning in Social Dilemmas}\label{subsec:rldilemmas}

Repeated versions of the games can be implemented as learning environments for RL agents. A social dilemma environment can be implemented for RL agents as a two-player iterated game, played over $T$ iterations \cite{littman1994markov}. This is summarised in Figure \ref{fig:diagram}. At each iteration, a moral player $M$ and an opponent $O$ play the one-shot matrix game corresponding to a classic social dilemma (see Table \ref{tab:three_games} for examples). At every time step $t$, the learning agent $M$ observes a state, which is the pair of actions played by $O$ and $M$ at the previous time step $t-1$: $s_M^{t}=(a_O^{t-1},a_M^{t-1})$, and chooses an action (Cooperate or Defect) simultaneously with the opponent $O$: $a^t_M, a^t_O \in \{C, D\}$. The player $M$ then receives a reward $R_M^{t+1}$ and observes a new state $s_M^{t+1}$. In traditional RL, this reward would be provided by the environment (i.e., \textit{extrinsic reward}), and in the social dilemma games it would correspond directly to the payoffs from the game (presented in Table \ref{tab:three_games}). 

\subsubsection{Moral Preferences as Intrinsic Rewards}\label{subsec:intrinsicrewards}

Intrinsic motivation is a concept originating in human Psychology \cite{deci2013intrinsic}, whereby humans' choices (such as the choice of a job) may not be associated with some external reward (such as money), but rather be driven by internal values (such as autonomy, excellence or intellectual curiosity, suggested by \shortciteR{wrzesniewski1997jobs}). In Philosophy, intrinsic motivation may align to the concept of will \cite{aristotle} or happiness \cite{Bentham1996}. 

\shortciteA{deci1999meta} and \citeA{Oudeyer2007intrinsic} draw interesting parallels between intrinsic motivation and rewards. The implication for RL is that effective learning can result from modifying the agent such that they take into consideration an internal reward signal rather than only rewards from the environment. Early successes with this approach were reported by \shortciteA{chentanez2004intrinsically} in curiosity-driven exploration, where curiosity was modelled as an intrinsic reward obtained when the agent visits unfamiliar states.

The intrinsic rewards framework is directly applicable to the idea of learning morality from experience. A small set of existing works in Multi-Agent RL modified the reward signal for agents in social dilemma games to embed an internal moral value or preference into reward functions \shortcite{Capraro2021,hughes2018inequity,mckee2020social,tennant2023modeling,tennant2024dynamicsmoralbehaviorheterogeneous}. In these works, moral preferences have either been expressed as an operation on the extrinsic rewards (applying some function over the player's own reward and that of their opponents), or a conditional function based on actions performed previously. A related line of work modifies reward functions to include an awareness of the opponents' learning \shortcite{foerster2018lola,jaques2019social} - but these are not strictly morality-focused implementations and thus lie beyond the scope of the studies reviewed here. 

\begin{figure}[t]
  \centering
  \includegraphics[width=0.8\linewidth]{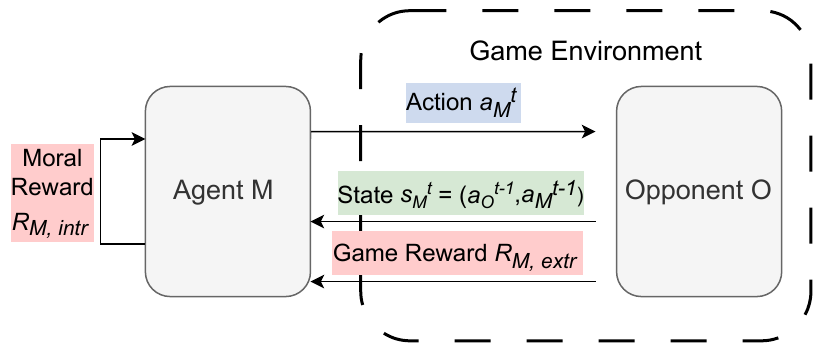} 
\caption{A step in a two-player game modelled as an RL process, from the point of view of a learning moral agent $M$ playing against a learning opponent $O$. Agent $M$ calculates their moral (intrinsic) reward according to their own ethical framework, which may or may not consider the game (extrinsic) reward coming from playing the dilemma.}
\label{fig:diagram}
\end{figure}

A visualisation of the intrinsic rewards approach is presented in Figure \ref{fig:diagram}. In this formulation, for player $M$, we contrast traditional agents that learn according to an \textit{extrinsic} game reward $R_{M_{extr}}$, which is simply the payoff associated with the joint actions $a_M^t,a_O^t$ (as defined in Table \ref{tab:three_games}), versus agents that learn according to an \textit{intrinsic} reward $R_{M_{intr}}$, based on some moral principle. In Section \ref{subsec:existingmoralframeworks}, we expand on the types of principles that can be encoded as intrinsic rewards.

In practice, both extrinsic and intrinsic rewards can be combined in a multi-objective manner \shortcite{vamplew2018human} to create an agent able to pursue multiple goals (for example, trying to obtain both moral reward and the reward from the task at hand - \shortciteR{ijcai2019p891}). This was the approach used by \shortciteA{chentanez2004intrinsically} to create their `curious' yet task-specific agent. Furthermore, multi-objective RL approaches can also be used to combine multiple intrinsic moral rewards to create a multi-faceted moral agent (e.g., see the multi-objective moral agent from \shortciteR{tennant2023modeling}), which allows for the creation of more complex and comprehensive moral agents.  

This case study implements a \textit{decentralised} RL approach where each agent chooses an action in the game without a central planner. A parallel line of work exists in the modelling and ethics literature, where an `ethical' agent is implemented as a centralised mechanism on the environment - this has applications in entirely different domains such as governmental policy design \shortcite{Koster2022democratic}, and lies beyond the scope of this work, which focuses on the moral alignment of individual agents.

\subsubsection{Formalising Existing Ethical Frameworks as Rewards for AI Agents}\label{subsec:existingmoralframeworks}

A variety of moral frameworks has been developed over the millennia in the fields of Moral Philosophy, Psychology, Biology and Economics. We review a representative set of these in Table \ref{tab:moral_definitions_sources}, formulating them as potential reward functions for agents in two-player games. A growing body of RL work demonstrates that mathematical formulations of these frameworks are able to effectively encourage agents to learn more prosocial and cooperative behaviour in social settings \shortcite{hughes2018inequity,mckee2020social,tennant2023modeling,tennant2024dynamicsmoralbehaviorheterogeneous}. 
\begin{table*}[ht!]
  \centering
\begin{tabular}
    {>{\raggedright}p{0.14\linewidth} 
    >{\raggedright}p{0.16\linewidth} 
    >{\raggedright}p{0.27\linewidth} 
    >{\raggedright}p{0.11\linewidth} 
    p{0.15\linewidth}}
\toprule
     \small\textit{Moral Agent Type } &  \small{\textit{References}} &  \small{\textit{Formulation in terms of reward}} &  \small{\textit{External / Internal}} & \small{\textit{Consequentialist / Norm-based}} \\ 
\hline
    Utilitarian or Altruistic & \small{\citeR{Bentham1996,andreoni2002giving,charness2002understanding,Ledyard1995inbook,Levine1998modeling} } &  \small{Sum of all players' payoffs, with various weights on on own vs. others' payoffs.} &  \small{\textit{either}} &  \small{Consequentialist} \\ 
\hline
    Inequity averse & \small{\citeR{fehrschmidt1999theory,fehr2004social} } &  \small{Difference between own payoff and others' payoffs, with parameters for advantageous vs disadvantageous inequity between them.} &  \small{Internal} &  \small{Consequentialist} \\
\hline
    Equality- or Fairness-preferring & \small{\citeR{gini1912variabilita,rapoport1974prisoner,Bolton2000ERC}} &  \small{Ratio (e.g., Gini coefficient) between own and opponent's payoffs; or a weight on the distance between own payoff vs fair share.} &  \small{Internal} &  \small{Consequentialist} \\ 
\hline
    Following a `Veil of Ignorance' & \small{\citeR{Harsanyi1961,Rawls1971}} &  \small{E.g., minimum payoff of any player (assuming an agent chose to distribute payoffs not knowing where they would end up in the distribution).}  &  \small{\textit{either}} &  \small{Consequentialist} \\
\hline
    Deontological & \small{\citeR{kant1981grounding,Capraro2021,kimbrough2023theory,krupka2013identifying,levitt2007laboratory}} &  \small{Positive (or negative) reward if the action taken adheres to (or violates) a pre-defined norm. Agent may enforce norm in others or not.} &  \small{External} &  \small{Norm-based} \\ 
\hline
    Virtue Ethics  \small{(can be based on other types above)} & \small{\citeR{aristotle,graham2013moral,rapoport1974prisoner} } &  \small{Reward for adhering to a single specific virtue (e.g., cooperativeness), or a mix of virtues (e.g., sum cooperativeness + equality rewards).} &  \small{Internal} &  \small{\textit{either or both}} \\
\bottomrule
\end{tabular}
\caption{Moral frameworks proposed in past literature in Philosophy, Psychology, Biology and Economics, and potential implementations in terms of rewards for RL agents.} 
\label{tab:moral_definitions_sources}
\end{table*}

We classify possible moral agent types in Table \ref{tab:moral_definitions_sources} into Consequentialist or Norm-based. \textit{Consequentialist} morality focuses on the outcomes or consequences of an action. In dilemma games, this can be easily implemented as an operation on the two players' payoffs, including objective functions that are Utilitarian \cite{Bentham1996}, defining actions as moral if they maximise total utility for all agents in a society, or Altruistic \cite{Levine1998modeling}, where an agent is inclined to maximise their opponent's winnings, perhaps without even considering their own. Consequentialist morality also includes fairness-based objective functions. These can be based on the difference between two players' rewards, including the inequity aversion models of \citeA{fehr2002social,fehr2004social}, or other calculations of the ratio between payoffs with different preferences for fair or equal outcomes \cite{Bolton2000ERC,gini1912variabilita,rapoport1974prisoner}. Finally, consequentialism also accounts for the principles of distributive justice of \citeA{Rawls1971}, in which agents distribute resources without knowing where they will end up in that distribution.

\textit{Norm}-based morality, on the other hand, focuses on defining actions as moral so long as they adhere to a society's external norms or to a specific agent's duties, regardless of the consequences further down the line. This framework includes the Deontological ethics of \citeA{kant1981grounding}, which identifies specific acts as morally required, permitted or forbidden. An example norm for dilemma game-playing agents may be conditional cooperation, in which an agent is expected to cooperate against a cooperative opponent \cite{fehr2004social}. Further norm-based objective functions have been formalised by \shortciteA{Capraro2021,kimbrough2023theory,krupka2013identifying} and \shortciteA{levitt2007laboratory}, who propose various weights on personal versus social importance of the norm at hand.

A third type of ethical framework that is often distinguished is \textit{Virtue Ethics} of \citeA{aristotle} - in this line of reasoning, moral agents act according to their certain internal virtues. In practice, these virtues themselves often have Consequentialist or Norm-based foundations - consider, for example, the virtues of fairness or care, respectively \shortcite{graham2009liberals,graham2013moral}. The key distinction of Virtue Ethics is that a single agent may rely on more than one type of virtue, and different agents may weigh the virtues differently against one another \shortcite{aristotle,graham2009liberals}, so a more expressive way of modelling virtue ethics might be through a weighted, multi-objective paradigm - such as that suggested by \shortciteA{vamplew2018human}.

In Table \ref{tab:moral_definitions_sources}, we additionally distinguish agents by the external or internal nature of their moral motivation - while each of these frameworks can be formalised as an intrinsic reward function for the RL agent, sometimes the moral aspect of the intrinsic motivation comes from outside the agent themselves. An example of this is where certain moral norms exist \textit{externally} in a society, so the agent prefers to adhere to them in their individual behaviour. 

The final step for formalising morality as intrinsic motivation for RL agents, then, is to formulate precise mathematical definitions of rewards based on these frameworks. Using the domain of social dilemmas introduced above, \shortciteA{tennant2023modeling} propose a formulation in terms of actions and/or outcomes at some time step $t$ in a two-player, two-action social dilemma game. For illustration, we present their mathematical definitions of five types of moral rewards in Table \ref{tab:moral_definitions} \shortcite{tennant2023modeling}. 

\begin{table}[h]
  \centering
  \begin{tabular}{ll}\toprule
    \textit{Moral Agent Type} & \textit{Moral Reward Function}  \\ \midrule
    Utilitarian  & $R_{M_{intr}}^t=R_{M_{extr}}^t + R_{O_{extr}}^t$ \\ 
    Deontological  & $R_{M_{intr}}^t= 
\begin{cases}
    $--$\xi,& \text{if } a_M^t=D ,  a_O^{t-1}=C \\ 
    0,              & \text{otherwise}
\end{cases}\ $ \\ 
    Virtue-equality & $R_{M_{intr}}^t=1-\frac{|R_{M_{extr}}^t-R_{O_{extr}}^t|}{R_{M_{extr}}^t+R_{O_{extr}}^t}$ \\ 
    Virtue-kindness  & $R_{M_{intr}}^t= 
\begin{cases}
    \xi,& \text{if } a_M^t=C  \\
    0,              & \text{otherwise}
\end{cases}\ $  \\ 

    Virtue-mixed  &
    $ R_{M_{intr}}^t= 
\begin{cases}
        \beta*(1-\frac{|R_{M_{extr}}^t-R_{O_{extr}}^t|}{R_{M_{extr}}^t+R_{O_{extr}}^t})
            + (1-\beta)*\hat{\xi}
    ,& \text{if } a_M^t=C  \\
        \beta*(1-\frac{|R_{M_{extr}}^t-R_{O_{extr}}^t|}{R_{M_{extr}^t}+R_{O_{extr}}^t})
    ,              & \text{otherwise}
\end{cases}
$ \\
    \bottomrule
    
  \end{tabular}
\caption{Definitions of a sub-set of intrinsic moral reward types, from the point of view of the moral agent $M$ playing a social dilemma game versus an opponent $O$.}
\label{tab:moral_definitions}
\end{table}

In this set, given two players $M$ and $O$, Consequentialist morality is implemented as operations on the two players' extrinsic rewards $R^t_{M_{extr}}$ and $R^t_{O_{extr}}$. Specifically, for a \textit{Utilitarian} agent, the reward is simply the sum of the two players' payoffs. For an equality-focused agent \textit{Virtue-equality}, the reward is the ratio of the two payoffs, based on the Gini coefficient \cite{gini1912variabilita}, transformed so that larger values mean greater equality between the two agents.

Norm-based morality, on the other hand, is operationalised in terms of a positive reward for an agent whose action $a^t_M$ adheres to a given norm, or a negative reward for actions violating a given norm. For example, a \textit{Deontological} agent following the norm of conditional cooperation (defined above) receives a negative reward of value $\xi$ for defecting against an opponent who previously cooperated. The other norm-based agent \textit{Virtue-kindness} follows a simple cooperative norm, obtaining a positive reward $\xi$ for cooperating, regardless of what their opponent did previously. 

Finally, these examples also outline one multi-objective reward function \textit{Virtue-mixed}, which linearly combines the two separate virtues `equality' and `kindness', with a weighting parameter $\beta$, and a parameter $\hat{\xi}$ used to fit the `kindness' element of the reward to the same scale as equality. 

\shortciteA{tennant2023modeling} implement the five moral intrinsic rewards defined in Table \ref{tab:moral_definitions} (as well as a Selfish baseline agent) and show that Q-learning agents learning against a variety of opponents are able to learn optimal moral policies given sufficient exploration. A further study by \shortciteA{tennant2024dynamicsmoralbehaviorheterogeneous} extends this analysis to populations involving a partner selection mechanism and a larger amount of possible prosocial and antisocial moral agent types. 

\subsection{Case Study: Fine-tuning LLM Agents with Intrinsic Rewards}\label{subsec:finetuneLLMspaper}

Finally, to complement the intrinsically moral RL agents described above, we consider the case of fine-tuning LLM-based agents for moral alignment. This work relies on the use of LLMs as decision-makers \shortcite{park2023generative,vezhnevets2023generative,wang2023voyager} in strategic scenarios \shortcite{gandhi2023strategicreasoninglanguagemodels,swanepoel2024artificial,zhang2024llmmastermindsurveystrategic}, often studied using game-theoretic environments such as social dilemma games discussed above \shortcite{akata2023playing,Fan2024LLMRationalPlayers}. Thus, the approach of fine-tuning LLMs with RL reconciles the previous implementations of hybrid moral alignment in pure-RL agents with the more general and currently popular domain of LLM agents.

As discussed previously, the predominant approach for aligning LLMs to human values is by training reward models on human preference data (i.e., RLHF), and fine-tuning an LLM using that reward model, though non-RL approaches, such as DPO \shortcite{rafailov2024dpo}, enable preference-based fine-tuning without the need for RL-based training. This approach teaches the LLM the values that are implicit in the human preference data, without any explicit definition of what they are. A core disadvantage of this, as discussed in Section \ref{sec:learningandmorality}, is the lack of transparency interpretability and control, and the risks of reward-hacking. 

\shortciteA{tennant2024moralLLMagents} instead propose a methodology for fine-tuning LLMs based on intrinsic rewards rather than human-generated preference data (as in the case of RLHF). The intrinsic rewards in this approach are defined in terms of actions or consequences in a specific environment, but the policies learned via such fine-tuning can, in theory, generalise to other environments. \shortciteA{tennant2024moralLLMagents} present a study fine-tuning LLM agents using the rich domain of social dilemmas described in Section \ref{subsec:environments}. For LLM agents in particular, decision-making scenarios or games can be expressed in the form of text, with specific tokens corresponding to each action choice. During fine-tuning, the model learns to associate certain action tokens and strategies with rewards, with online RL implemented via on-policy methods such as PPO (\shortciteR{schulman2017PPO}, though offline implementations are also possible, see \shortciteR{snell2023ILQL}). 

The results in \shortciteA{tennant2024moralLLMagents} offer some evidence of the ability of LLM agents to generalise moral policies learned in one game to other environments characterised by a similar game structure - specifically, from the IPD agent to four other 2x2 matrix games. The challenges in this implementation include the fact that fine-tuning on a small set of specific action tokens may make the models over-fit to those tokens - indeed, \shortciteA{tennant2024moralLLMagents} observed that the fine-tuned model was more likely to produce the action tokens in response to unrelated questions of a similar format. Nevertheless, overall, this work suggests an interesting emerging direction in using intrinsic rewards as a hybrid and more controllable and interpretable implementation of moral alignment in agentic LLM systems.

\section{Evaluating Moral Learning Agents}\label{sec:evaluation}

\subsection{Outcomes, Behaviours and Qualitative Assessment}\label{subsec:evaluation-overview}

After training such morally motivated agents, researchers must evaluate whether their learning was ethically effective. This requires the definition of a set of \textit{outcome metrics}. Traditional Reinforcement Learning research would train agents in a specific simulated environment and measure some population-level outcome such as cumulative reward. However, as discussed in Section \ref{subsec:bottomup}, further evaluation metrics must be designed to identify potential reward-hacking behaviour of RL agents - in other words, one must also measure concepts or behaviours that were not directly encoded in the agents' reward functions. Ideally, these measurements should be broad enough to cover a vast number of diverse ethical considerations in a wide range of scenarios. The challenges of this have been discussed by \shortciteA{Reinecke2023puzzle}. As an illustration, in Section \ref{subsec:evaluation-outcomes} below, we present an example set of population-level outcome metrics that were proposed by \shortciteA{tennant2023modeling} for the case of social dilemma games. To complement this example of measuring RL agents in  social dilemma environments, we also present a discussion of how LLM-based agents can be measured in terms of behavioural alignment (see Section \ref{subsec:evaluation-llms}).

On top of population-level outcomes, a game-theoretic evaluation of emerging \textit{behaviours} may also be of interest, evaluating actions taken in the game over the course of learning, and the extent to which these actions are aligned with moral norms. In social dilemmas, this must include a measurement of cooperation displayed by each agent towards the end of the learning period as well as the emerging strategies (i.e., state-action combinations chosen by the agent). This allows for a non-consequentialist evaluation of learned behaviours and policies, and provides better insight for norm-based morality. 

Finally, an alternative means of evaluation could include a \textit{qualitative assessment} by a human who interacts with an agent. Within the domain of autonomous vehicles, large-scale evaluations of hypothetical agent behaviours have been conducted by \shortciteA{awad2018moral} using survey studies, and focusing on the trolley problem \cite{trolleyproblem} in particular as an agent environment. Much more qualitative evaluation is required in other domains and sets of problems to understand human perception of AI-assisted decision making more generally.

\subsection{Example: Measuring Outcomes in Social Dilemma Environments}\label{subsec:evaluation-outcomes}

In the dilemma scenarios presented in Section \ref{subsec:environments}, the choice of one agent affects the outcomes of another. Thus, it is possible that scenarios will arise in which agents intentionally designed with a certain morality (e.g., maximise `collective reward') actually display behaviour that maximises their desired outcome at a cost of some other important outcome (e.g., `equality'). For this purpose, on top of analysing individual rewards, researchers have highlighted the importance of measuring a range of population-level outcomes \shortcite{hughes2018inequity,leib02017multiagent,tennant2023modeling}. 

The most popular social outcome metric is the collective payoff for all agents - used by \shortciteA{leib02017multiagent,hughes2018inequity,mckee2020social}. Additional moral evaluation metrics can include a measure of equality such as the Gini coefficient \cite{gini1912variabilita}, and an egalitarian measure of the minimum payoff obtained by any agent in the population on every iteration \cite{Rawls1971}. These three social outcome metrics were formalised in \shortciteA{tennant2023modeling} as a cumulative return $G$ (i.e., the total reward accumulated over all the $T$ iterations of an iterated game) for both players $M$ and $O$ as follows:

\begin{align} 
G_{collective}&=\sum_{t=0}^{T} {(R_{M_{extr}}^t+R_{O_{extr}}^t)} \\
G_{Gini}&=\sum_{t=0}^{T}{(1-\frac{|R_{M_{extr}}^t-R_{O_{extr}}^t|}{R_{M_{extr}}^t+R_{O_{extr}}^t})} \\
G_{min}&=\sum_{t=0}^{T}{\min(R_{M_{extr}}^t,R_{O_{extr}}^t)} . \end{align} 

In \shortciteA{tennant2023modeling}, the five agents from Table \ref{tab:moral_definitions} are systematically evaluated on three distinct social dilemma games in terms of the cumulative outcomes defined above and cooperation levels and strategies. The results show that - with sufficient exploration and learning time - all types of moral agents are able to learn fully cooperative policies, though the `equality' definition produces the least efficient learning agent, which is reflected in the social outcome metrics. 

\subsection{Example: Measuring Moral Behaviour in Language Models}\label{subsec:evaluation-llms}

The domain of Large Language Models (LLMs) provides a different set of challenges for measurement. Its inputs and outputs are open-ended, such that the possible action space for a single generated token is as large as the vocabulary size associated with the model. Furthermore, the users of the model are likely to have diverse moral preferences of its users (discussed above). This makes it harder to define a standard set of metrics to evaluate the moral learning of LLMs. Nevertheless, a large set of benchmark metrics and data sets have already been proposed for measuring morality in these systems (see \shortciteR{hendrycks2021ethics,hendrycks2021would,pan2023rewards,Reinecke2023puzzle}). Capitalising on the open-ended nature of human language, studies such as \shortciteA{tennant2024moralLLMagents} have also assessed the generalisation of moral policies from one environment to others, though more remains to be done in terms of assessing their performance in real-world scenarios.

\section{Outlook \& Implications}\label{subsec:implications}

Our discussion suggests that the traditional fully top-down (e.g., rule-based) methods, or the currently popular fully bottom-up approaches to AI morality and safety (e.g., fine tuning of LLMs based on Reinforcement Learning from Human Feedback), whilst intuitively attractive because of the control they seem to impose (top-down), or the generality they may offer (bottom-up), are limited and pose potential risks for systems deployed in the real world. We motivate the use of a combination of interpretable top-down quantitative definitions of moral objectives, based on existing frameworks in fields such as Moral Philosophy, with the bottom-up advantages of trial-and-error learning from experience via RL. We argue that this hybrid methodology provides a powerful way of studying and imposing control on AI systems while enabling adaptation to dynamic environments. We review four case studies combining moral principles with learning, which provide proof-of-concept for the potential of this hybrid approach in creating more prosocial and cooperative agents. We hope that the unifying framework defined in this paper helps future practitioners and policymakers identify new possible avenues for future work. 

We appreciate the fact that quantifying morality by translating complex moral frameworks into elementary reward functions, learning constraints or textual principles in this way is certainly a simplification of reality. Nevertheless, the original moral philosophical frameworks discussed in this work might be seen as a form of abstraction and, in a sense, simplification of the complex and multi-faceted human condition, and yet these have served the field for centuries. 

We believe there is a need to study these hybrid approaches in a variety of social environments for morally-informed learning, including other social dilemma games and games involving a large number of players. The specific moral reward functions proposed in Tables \ref{tab:moral_definitions_sources} and \ref{tab:moral_definitions} and in a number of existing practical works \shortcite{Capraro2021,hughes2018inequity,mckee2020social,tennant2023modeling}, as well the other methodologies reviewed \shortcite{bai2022constitutional,Neufeld2022enforcing}, are a starting point for building more sophisticated moral agents, including those based on more than one moral objective \shortcite{vamplew2018human}. In general, the reward-based methods have a disadvantage of the fact that they rely on a clear scalar payoff structure, and in the case of some of the intrinsic moral rewards proposed - a discrete set of actions. An open research question remains about how to obtain this reward structure in open-ended tasks while still maintaining principled control over the morality implemented in the reward.

We hope that, in the future, researchers can build on these ideas and further investigate the effectiveness of each of the proposed moral frameworks experimentally. As discussed, the moral preferences of the target audience of any AI system will vary depending on their cultural and political norms \shortcite{graham2013moral} - in this paper, we did not aim to promote any one moral framework, but instead demonstrated the ways in which different moralities can be implemented using the same methodology of intrinsic rewards. 

Finally, it is worth noting that approaches described above can bring benefit back to the disciplines they are founded on and took inspiration from. In particular, the proposed methodology can be used to reflect upon and understand different moral frameworks and their impacts, especially in complex situations. For example, it can be used for simulating human societies, studying the outcomes deriving from the presence of agents with specific moral preferences, or to study emergent behaviours in societies composed of agents with different moral principles. In general, this can be seen as a practical example of `Computational Philosophy' \shortcite{MayoWilsonConor2021Tcps}. Moreover, in general terms, modelling moral behaviour as a reward-maximising $\epsilon$-greedy choice made by a continually learning agent can provide insights for Evolutionary Game Theory. Our simulations with learning agents can discover how certain moral behaviours can evolve (and might have evolved) in a society given a payoff structure and different distributions of preferences and norms within a population. Additionally, an analysis of the learning dynamics of agents can provide insights for Developmental Psychology, possibly offering a way to test various theories of moral development in children \cite{gilligan1982,gopnik2009philosophicalbaby,kohlberg1975}. 


\section{Conclusion}

In this paper, we have made the case for the use of learning with explicitly defined moral principles for developing morality in AI agents, contrasting it with more traditional top-down implementations in AI safety and ethics, and the more recent entirely bottom-up methods for learning implicit values from human feedback, such as those used for fine-tuning LLMs. We have suggested RL as one effective mechanism for learning morality from experience in a given environment, and reviewed a set of case studies
to analyse different moral alignment approaches that have been implemented for agentic systems in practice. In two of these case studies (based on the pure-RL or LLM domains), we have outlined a hybrid methodology in which agents learn morality via intrinsic rewards based on classic moral frameworks, and presented a formalisation of a set of specific reward functions using social dilemma environments. We have also reviewed solutions based on alternative approaches, such as constrained RL or fine-tuning LLMs from constitution-based feedback. Finally, we have proposed an example of potential evaluation of the moral outcomes and behaviours emerging from interactions between different agents over time, and outlined the existing open challenged in this area. 

We hope that this article can motivate further work in the area of developing moral AI agents via learning with the help of top-down explicit moral principles (i.e., in  a hybrid way). While there are substantial challenges in implementing the proposed design guidelines in practice, we believe it provides a promising avenue for addressing the growing safety concerns around emerging advanced AI.

\newpage 
\section*{Definitions}\label{sec:definitions}

\begin{longtable}{>{\raggedright}p{0.18\linewidth} | p{0.77\linewidth}}
    \toprule
    \textbf{Concept} & \textbf{Definition} \\ \midrule
    Agent & An entity able to take actions / make decisions or choices in an environment. \\ 
 \hline
    Bottom-up & Development of a system / inference of values through experience, organically, without planning the general architecture / constraints / components in advance. An example is Reinforcement Learning of desired behaviours from human demonstrations (e.g., via Inverse RL or RLHF). \\ 
 \hline
    Extrinsic Reward & A reward signal provided by the environment to the agent. For example, in a Prisoner's Dilemma, the game payoff for a player would constitute the extrinsic reward.  \\ \hline
    Intrinsic Reward  & A reward signal based on an internal value, preference or principle of the agent, rather than received from the environment. For example, in humans, the intellectual enjoyment in a job is an intrinsic `reward', while the associated pay is the extrinsic `reward'. In AI agents, intrinsic rewards can be based on a moral \textit{preference} for how group rewards should be distributed, or a moral \textit{norm} defining how agents should and should not act in certain situations (see below).  \\ \hline
    Moral framework &  [In Philosophy] A set of principles for how humans / agents should behave morally in a society (e.g., Utilitarianism, Deontological Ethics, Virtue Ethics). \\ \hline
    Moral norm & A norm in society about what constitutes moral behaviour. An example is `not defecting against a cooperator'. \\ \hline
    Moral value or preference & An internal representation of a moral principle. An example is a preference for a certain way of distributing resources (i.e., `fairness'), or a motivation to `care' for others. \\ \hline
    Social Dilemma & A situation where there is a trade-off between short-term individual reward and long-term collective outcomes. An example is the two-player matrix dilemma game with two possible actions (Cooperate or Defect), where players choose actions simultaneously without the ability to communicate, and receive payoffs based on the joint action. \\ \hline 
    Top-down & Development of a system by defining / planning the architecture and explicitly defining all components, requirements and constraints in advance. An example is a system with moral rules based on deontic logic. 
    \\   
    \bottomrule

\end{longtable}


\appendix





\section*{Appendix A. Other Games Involving Morality}\label{appendix:a}

Other types of games in which it is possible to model human morality are sequential games \shortcite{Harsanyi1961,leib02017multiagent,jaques2019social}, games with more than two players \shortcite{Liebrand1983,HardinG1968Totc}, and games with incomplete information, first introduced by \shortciteA{Harsanyi1995incompleteinfo}. 

Sequential games model social situations in which one player takes an action first, then the other chooses how to respond given how the first player acted. An example of this is the Ultimatum game \shortcite{Harsanyi1961}, in which the first player offers a way to split a sum of money, and the second player either accepts the split or rejects it, in which case neither player gets anything. Here the first player may be incentivised to keep a larger sum for themselves, but they risk pushing the opponent too far and ending up with nothing. Applied to ethics, the Ultimatum game can model the principle of fairness in both players (fairness in one's own behaviour for the proposer, and punishment for unfair behaviour by the receiver), providing an interesting test bed for moral learning agents. 

Games with more than two players are an extension of the dyadic dilemma games presented in the previous section, but they involve an entire population playing at once. The Prisoner's Dilemma can be extended to a social version, for example by considering the problem of pollution and how each individual contributes to it in a society \shortcite{Liebrand1983}. The Volunteer's Dilemma can also easily be extended to a case in which the society (and every player in it) benefits from someone else volunteering in their community, but each player is individually motivated to avoid volunteering themselves. Collectively, this type of situation is known as the Tragedy of the Commons \shortcite{HardinG1968Totc}. More recently, environments have been created specifically for training RL agents in such social dilemmas with spatial and temporal resolution, for example in the games Gathering \shortcite{leib02017multiagent} and Cleanup \shortcite{jaques2019social}. Multi-player dilemmas can introduce an interesting aspect of partial observability (i.e., incomplete information), as one agent may not be certain about what action another agent took previously just by observing the payoffs. This creates a layer of anonymity that can influence the development of certain moral norms in a society.

\vskip 0.2in
\bibliography{article}
\bibliographystyle{theapa}

\end{document}